# Analyzing Search Spaces with Wang-Landau Sampling and Slow Adaptive Walks

Susan Khor

Canada

slc.khor@gmail.com

## ABSTRACT

Two complementary techniques for analyzing search spaces are proposed: (i) an algorithm to detect search points with potential to be local optima; and (ii) a slightly adjusted Wang-Landau sampling algorithm to explore larger search spaces. The detection algorithm assumes that local optima are points which are easier to reach and harder to leave by a slow adaptive walker. A slow adaptive walker moves to a nearest fitter point. Thus, points with larger outgoing step sizes relative to incoming step sizes are marked using the local optima score formulae as potential local optima points (PLOPs). Defining local optima in these more general terms allows their detection within the closure of a subset of a search space, and the sampling of a search space unshackled by a particular move set. Tests are done with NK and HIFF problems to confirm that PLOPs detected in the manner proposed retain characteristics of local optima, and that the adjusted Wang-Landau samples are more representative of the search space than samples produced by choosing points uniformly at random.

## 1. INTRODUCTION

The difficulty of a search problem has been linked to the number of local optima and their placement in a search space relative to each other. Local optima may be clumped together forming deep-valleys or massive centrals [8]; or spread out as in the Schwefel function such that the fitness landscape approximates a unimodal only at a very high level of coarse-graining [9, 10]. Hence, numerous search space analysis techniques and even some stochastic search algorithms require knowledge of the local optima in a search space and their pair-wise saddle points or barrier heights. For example, local optima found at various temperatures make up the nodes of disconnectivity graphs [1]. The leaf nodes of barrier trees [4] are local optima found at a specific temperature or using a move set, and the non-leaf nodes are saddle points at which fitness barriers are minimal. Local optima form the nodes of inherent networks [2] and of Local Optima Networks [11]. The links in these networks depend on the detection of a transition state or saddle point between a pair of local optima. Stochastic search methods such as the hybrid genetic algorithm [5], the convex global underestimator [6], and basin hopping global optimization [18] whittle down a search space to its local optima or transform the energy function to associate a search point with a local minimum, and then focus the search effort on the reduced or transformed space.

But search spaces grow exponentially with problem size which places a severe space and time requirement on an enumeration approach of the entire search space. To address this, approaches involving walks, path sampling and fitness interval restrictions have been used to explore a search space. One of the earlier approaches employed steepest descent path to map configurations found through molecular dynamics simulation to the nearest local minimum [14]. Becker and Karplus use this method to construct their disconnectivity graph [1]. Steepest descent or gradient walks also appear in studies of combinatorial search spaces. The terminating configuration of such walks is considered a local optimum. Fonlupt et al compared results obtained by steepest descent walk to those obtained by random descent walk on the travelling salesman problem, and found no difference in the distribution of the quality (fitness) of the local optima, or in the characterization of the basins of attraction as highly-intertwined crater-like structures [5].

It is common for sampling techniques to rely on the selection of random search points to begin walks (e.g. [5]) which may or may not traverse hard to reach areas of a search space. To calculate their dispersion metric, Lunacek and Whitley use the fittest $n$ points in a set of points sampled uniformly at random from a search space to catch points likely to be locally optimal [10]. Also, restricting the construction of a barrier tree to a fitness range [7] may or may not reveal interesting areas of a fitness landscape. With Wang-Landau sampling [19] we avoid the use of such techniques, and gain some level of assurance that all regions of a search space are sampled. If you will, to touch all the different parts of an elephant, except in the case of search spaces, all fitness bins.

This is the final version which would have appeared in *GECCO'12 Companion*, July 7–11, 2012, Philadelphia, PA, USA as part of a workshop but was withdrawn due to funding problems.

## 2. PLOPs DETECTION

*Potential local optima points* (PLOPs) are points in a search space with high potential to be local optima. The algorithm to detect PLOPs relies on the notion of a *slow adaptive walker* which moves to a not already visited nearest fitter point until unable to do so. With no restriction placed on the maximum step size, all slow adaptive walks can in principle terminate at a global optimum. Intuitively, PLOPs are easy for a slow adaptive walker to reach but difficult to leave. A point is easy to reach if only a small step is required to move towards it from a less fit point. A point is difficult to leave if a large step is required to move from it to a fitter point. Hence steps from PLOPs are expected to be relatively larger than steps to PLOPs. We quantify this idea with the *local optima score* (*los*) defined in Fig. 1. Points with *los* > 0.0 are PLOPs.

```
if (in_max_stepsize == 0) los = 0.0
else
  if (out_min_stepsize == 0)
    los = 7.0  //arbitrary positive number
  else
    los = 0.0
    if (out_mode_stepsize > in_mode_stepsize)
      los +=(out_mode_stepsize - in_mode_stepsize)
    if (out_avg_stepsize > in_avg_stepsize)
      los += (out_avg_stepsize - in_avg_stepsize)
    if (out_min_stepsize > in_max_stepsize)
      los += (out_min_stepsize - in_max_stepsize)
```

**Fig. 1** *Local optima score* (*los*) calculation for each point. `in_max`, `in_avg` and `in_mode` respectively are the maximum, average and most frequently occurring step size in the set of incoming steps for a point. `out_min`, `out_avg` and `out_mode` respectively are the minimum, average and most frequently occurring step size in the set of outgoing steps for a point. When an exhaustive search for a next nearest fitter point is made, `out_mode_stepsize = out_avg_stepsize = out_min_stepsize`.

## 2.1 Algorithm

The PLOPs detection algorithm is outlined in Fig. 2. The set of input points P may be the complete enumeration or a sample of a search space. The ability to handle partial search spaces is one of the advantages of the approach to local optima detection proposed here. Since the definition of local optima is no longer tied to a specific move set, the sampling process can be decoupled from the move set. This opens up the possibility of using general sampling techniques, and preserving the properties of resultant samples, e.g. fitness distribution or density of states.

```
INPUT: A set of points P and their respective fitness values.
1. Take a slow adaptive walk from each point, and record the size of steps taken in each walk.
2. Calculate local optima score (los) for each point (Fig. 1).
OUTPUT: A set of PLOPs, i.e. points with los > 0.0
```

**Fig. 2** Outline of the PLOPs detection algorithm

Finding a nearest fitter point is computationally expensive. To reduce this cost when using exhaustive search, we use the dynamic programming technique and load the algorithm with pre-calculated sets of nearest fitter points, one for each point in P, i.e. the neighborhood of each point. Hence, finding a nearest fitter point for a point *p* is a matter of choosing a point uniformly at random from the neighborhood of *p*.

## 2.2 Testing Method

The purpose of the test is to assess how well the PLOPs detection algorithm detects local optima. We accomplish this by taking the *plef* score of each PLOP. The *plef* score for a point is the fraction of less or equally fit points in the 1-bit flip neighborhood of the point. By definition, a local optimum is not less fit than any of its neighbors. Thus points with *plef* = 1.0 are local optima (in the most restricted sense, i.e. 1 bit-flip move).

The *plef* test is conducted in the first instance on NK problems with random neighbourhood interactions [8]. We use N=16 with K= 4, 8 and 12. The problem size is small enough for our computer resources to enumerate the search space, without it being trivial. Points are binary strings of length N. Distances between points are in Hamming distance. Since NK problems rely on random values for fitness evaluation, 30 independent instances (both neighborhood and fitness values were randomized) were generated for each NK problem, and test results are summarized over 30 instances per NK problem.

We performed the *plef* test on the NK problems under three conditions: (i) ENUM where the set of input points comprise the entire search space, i.e. $2^N$ binary strings; (ii) AWL where the set of input points comprise points selected by Wang-Landau sampling (section 2.3); and (iii) RAND where the set of input points comprise points selected uniformly at random from the entire search space. An AWL sample is generated for each of the 30 instances per NK problem. RAND sample sizes match those of corresponding AWL samples. For convenience, we refer to the population of $2^N$ binary strings as the ENUM sample.

## 2.3 Search Space Sampling

The Wang-Landau algorithm[1] does a random walk and is biased to accept moves that visit areas of a search space (partitioned into fitness bins) which hitherto have been less explored.

An AWL sample is the set of points visited in one run of the Wang-Landau algorithm with the following typical parameter values: (i) the modification factor *f* starts at *e* and is reduced as $f_{t+1} = f_t^{0.5}$, (ii) epsilon is $10^{-8}$, (iii) the histogram is considered flat if every fitness bin has been visited at least 0.85 of the number of visits averaged over all bins, and (iv) the move operator is 1 bit-flip. Because we are not interested in the density of states per se and actually want to avoid ergodicity (visiting all points in the search space), and to exert some control over sample sizes, we adjust the Wang-Landau algorithm with the following additional terminating conditions: (i) the algorithm terminates when the histogram has been flat at least 5 times and the sample size is larger than the required minimum, or when (ii) the sample size is or exceeds the specified maximum. Presently, we set the maximum sample size to $2^N$ for $N \leq 16$ and $2^{16}$ otherwise[2]. The minimum sample size is half of the maximum sample size. Admittedly, these parameter values are somewhat arbitrary and calls for further study.

Fig. 3 traces the increase in sample size for three AWL sampling runs. Run #27 for NK(16, 4) demonstrates an instance where our parameter settings did not work well. AWL sampling ends up covering the entire search space because of the minimum sample size requirement. However, this is more the exception than the rule (hence the large variation for NK(16, 4) in Fig. 4). AWL samples cover on average about 60% of the search space for the NK problems. Fig. 4 shows the time taken by the Adjusted Wang-Landau algorithm, in terms of the average number of fitness function evaluations made, to produce the AWL samples for the NK problems, and the average AWL sample size as a proportion of the search space.

---

[1] Pseudo-code available at http://en.wikipedia.org/wiki/Wang_and_Landau_algorithm

[2] The upper limit on sample size for N > 16 could be increased with the use of more efficient techniques to find fitter nearest strings.

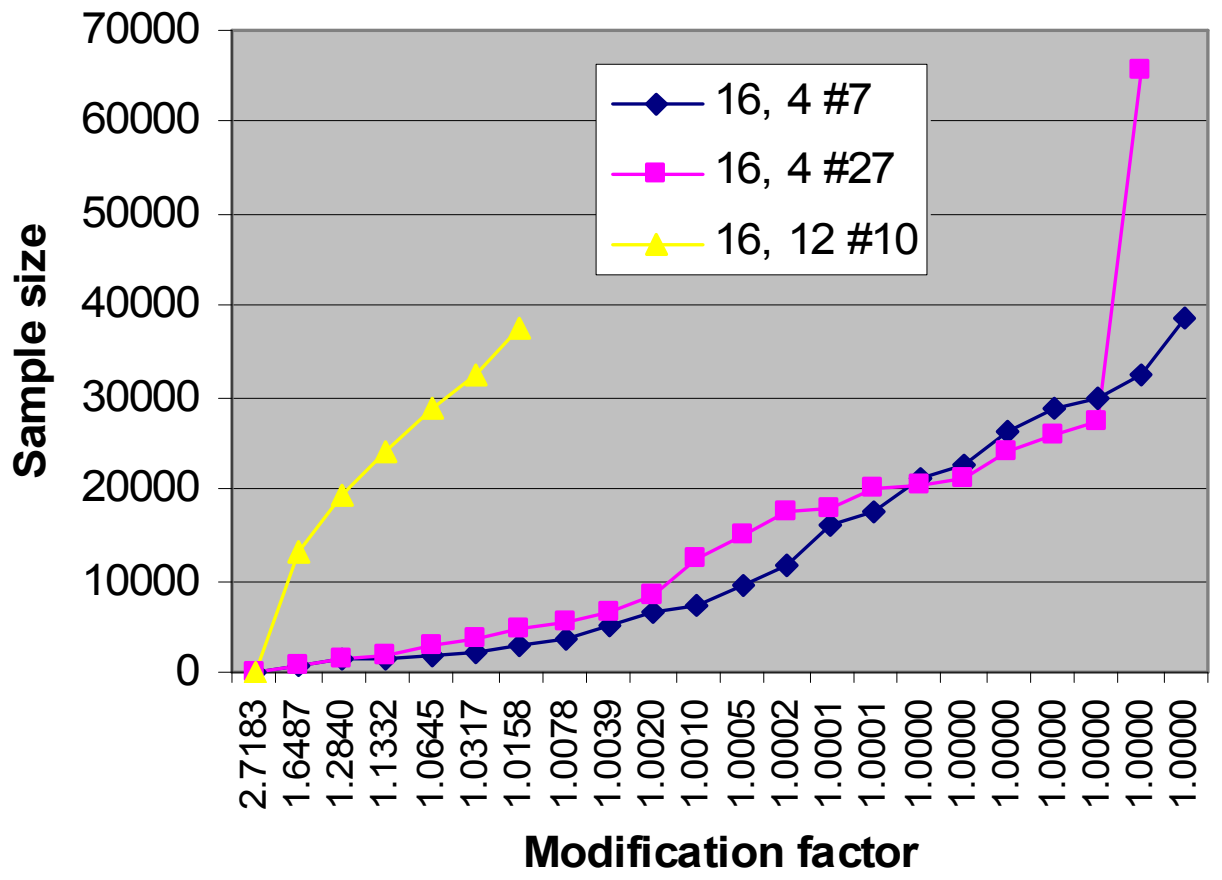

**Fig. 3 Sample sizes over iterations of the AWL sampling algorithm. Each time the histogram is flat, the modification factor is adjusted by taking the square root. Modification factors are shown to four decimal places.**

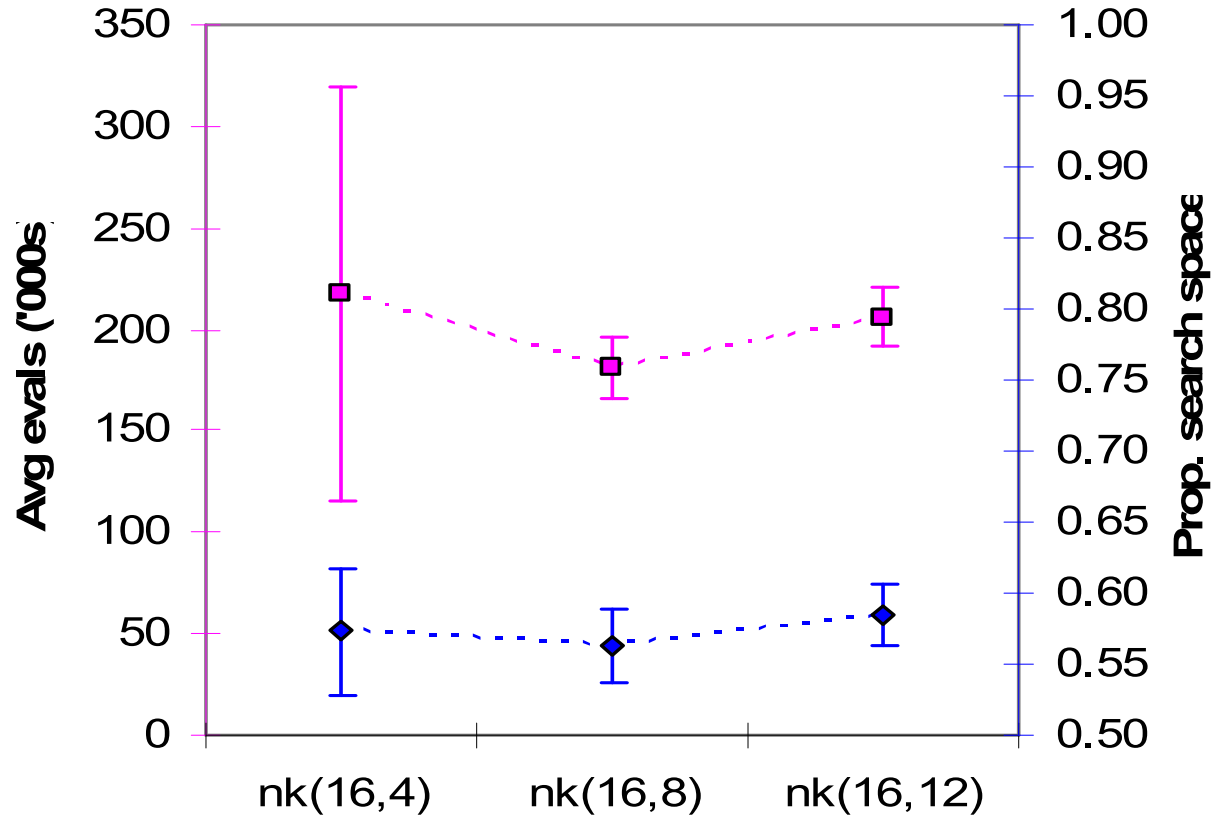

**Fig. 4 Average number of fitness function evaluations taken to generate AWL samples (left y-axis). AWL sample size as a proportion of search space (right y-axis). Error bars indicate the 95% confidence interval for the average.**

The average minimum and average maximum fitness values of AWL samples did not differ significantly (95%) from those of ENUM. The RAND samples tend to have slightly narrower fitness ranges. The strongest evidence supporting the use of AWL sampling over RAND is given in Fig. 5. It shows that AWL samples contain significantly more locally optimal points (points with *plef* = 1.0) than RAND samples. This occurs despite RAND sampling over 50% of the search space, and the number of local optima increasing with larger K for NK problems.

There is some difficulty with AWL sampling on problems like NK which do not have easily computed minimum and maximum fitness values (i.e. without enumerating the whole search space). Also, we found there may be gaps in fitness values for NK problems which need to be addressed when formulating the fitness bins. These practical issues of the Wang-Landau algorithm are known. Our present approach (which is feasible because N=16, and sufficient for our proof of concept purpose here) is to create fitness bins between the minimum and maximum fitness in increments of 0.1 and then remove bins which are impossible to fill. More dynamic approaches (e.g. [16]) could be applied in the future.

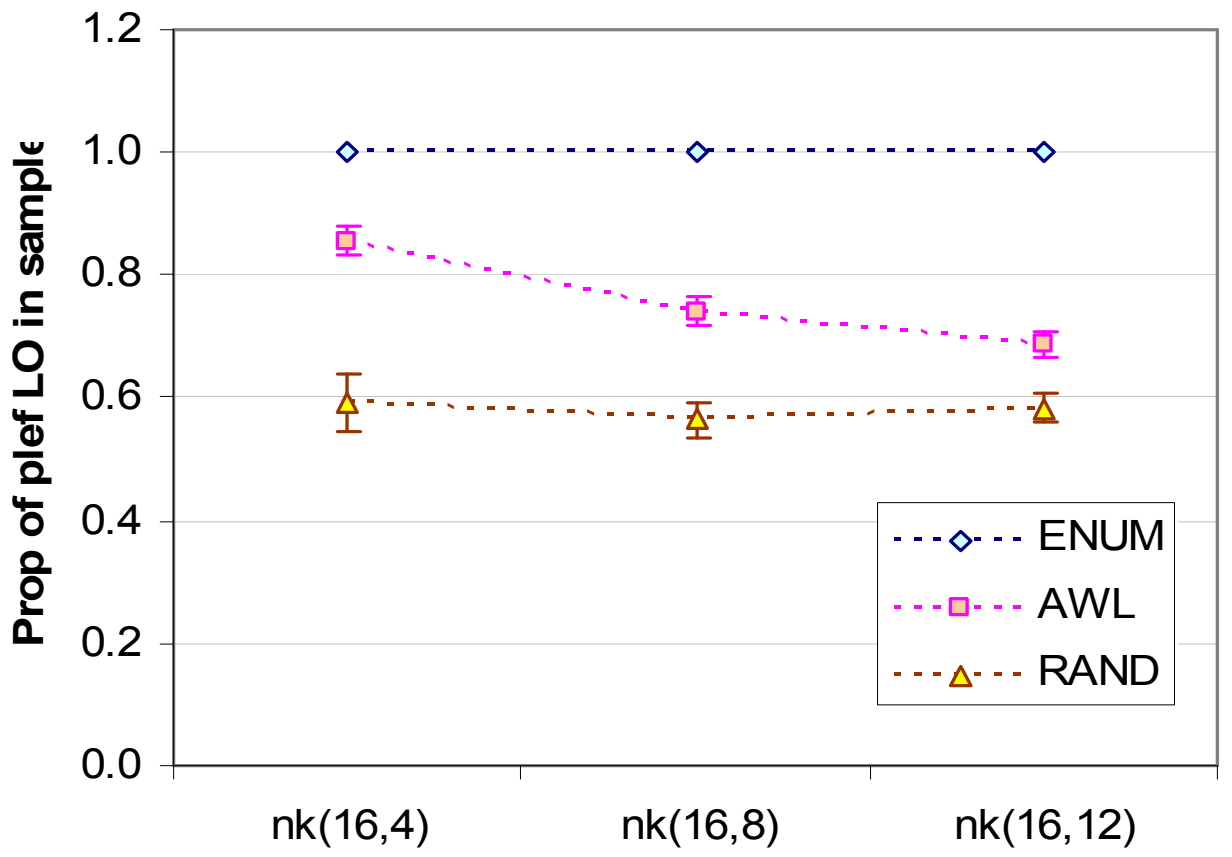

**Fig. 5 Proportion of *plef* LO (points with *plef* = 1.0) from the entire search space caught in a sample averaged over all samples/instances per NK problem. Error bars indicate the 95% confidence interval for the average.**

## 2.4 Results

How well does the PLOPs detection algorithm work? When a locally optimum point is present in a sample, it is detected over 94% of the time (Fig. 6). This result holds regardless of whether the slow adaptive walks are made over the entire search space (ENUM), or a subset of the search space (AWL and RAND). In this sense, the PLOPs detection algorithm works well. Not surprisingly, detection rate is highest when information used to compute los come from slow adaptive walks over the entire search space (ENUM). AWL samples have significantly higher detection rates than RAND samples (Fig. 6). This is further evidence supporting the use of AWL sampling over uniformly at random sampling.

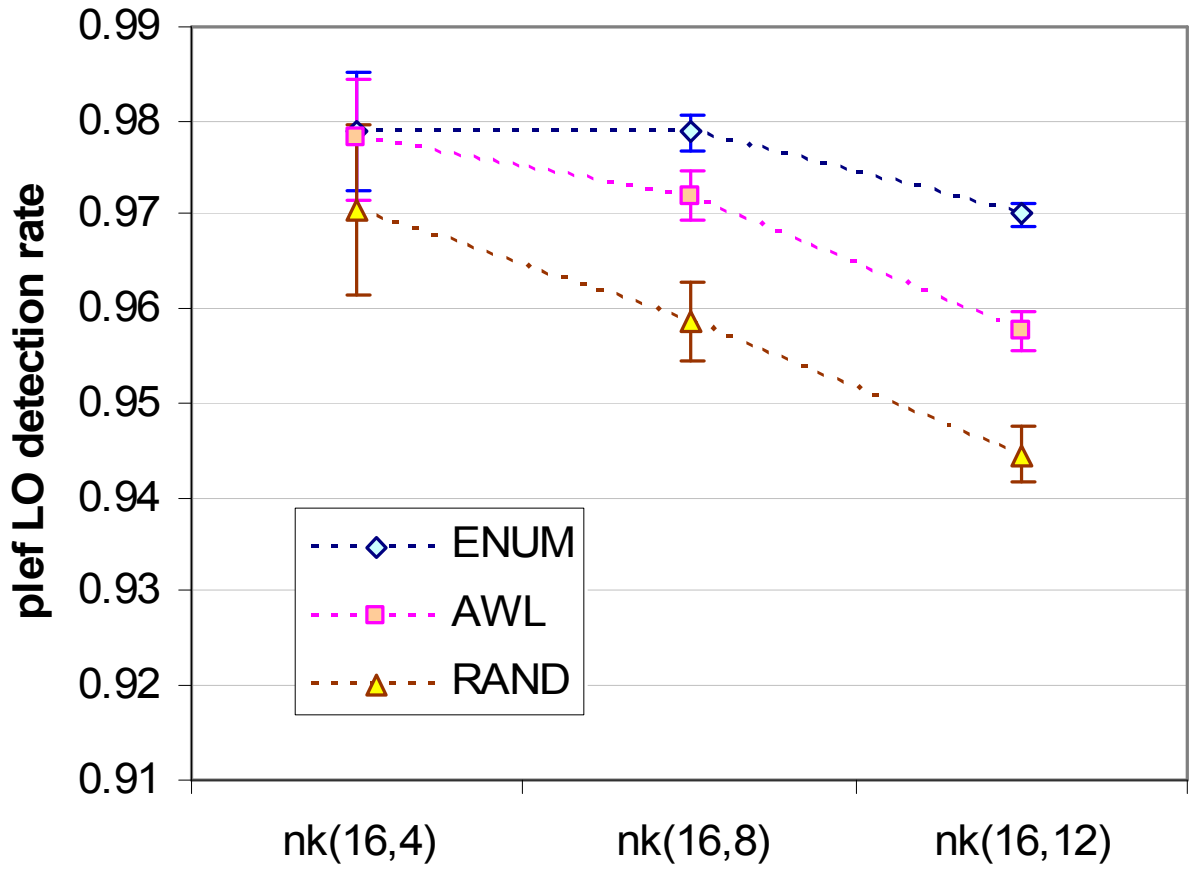

**Fig. 6 Proportion of *plef* LO (points with *plef* = 1.0) in a sample which are classified as PLOPs averaged over all samples/instances per NK problem. Error bars indicate the 95% confidence interval for the average.**

Fig. 7 compares the average number of PLOPs detected under the three test circumstances. We first note that in all three plots, the average number of PLOPs detected increases with increases in K, which is the expected pattern for NK problems. There is no significant difference between ENUM and PLEF (the actual number of points in the entire search space with *plef* = 1.0). However, the algorithm generates a significant number of false positives for AWL and RAND. This overestimation problem is worst for RAND (more evidence supporting the use of AWL sampling), and for smaller K (Table 1). Nonetheless, overestimation of PLOPs decreases with increases in K for both AWL and RAND possibly due to increasing number of local optima (when there are more targets, it's more difficult to miss).

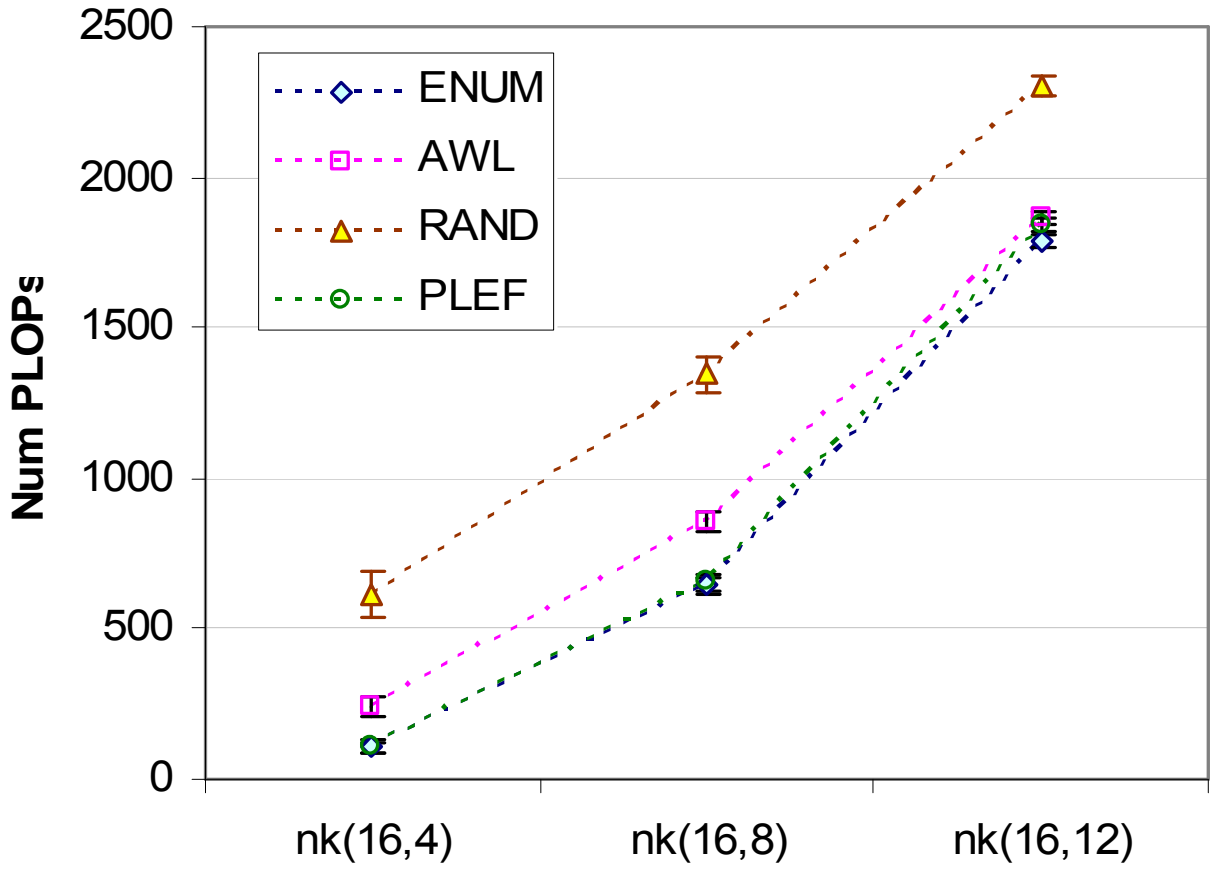

**Fig. 7 Number of PLOPs detected in a sample averaged over all samples/instances per NK problem. Error bars indicate the 95% confidence interval for the average. PLEF is the actual number of points in the entire search space with *plef* = 1.0 averaged over all instances per NK problem. Values plotted in this figure are used to calculate the overestimation factor in Table 1.**

**Table 1 PLOP overestimation factor: Average number of PLOPs / average number of *plef* LO**

| NK | ENUM | AWL | RAND |
|---|---|---|---|
| (16, 4) | 0.98 | 2.25 | 5.73 |
| (16, 8) | 0.98 | 1.30 | 2.05 |
| (16, 2) | 0.97 | 1.01 | 1.25 |

What of the quality (*plef* value) of the detected PLOPs, including the false positives? On average, PLOPs registered significantly higher *plef* values than their samples. ENUM PLOPs, because of their high *plef* LO detection rate and low overestimation factor, report average *plef* scores close to 1.0. AWL PLOPs fared better than RAND PLOPs with significantly higher average *plef* scores. PLOPs are also fitter on average than the sample they come from. So while the algorithm overestimates the number of PLOPs actually present in a sample, the points it selects as PLOPs share attributes of local optimum points, namely above average *plef* score and fitness.

# 3. CHARACTERISTICS OF PLOPs

In section 2.4, we demonstrated two characteristics of PLOPs which agree with standard literature on local optima in NK landscapes: (i) their numbers increase with K, and (ii) they have above average fitness. In this section, we extend this line of analysis to further evaluate the use of AWL sampling and the PLOP detection algorithm.

## 3.1 Basin of Attraction

The basin of attraction for a local optimum $x$ encompasses the set of points B($x$) such that a gradient walk from a point in B($x$) terminates at $x$. With this definition, it is possible for certain fitness landscapes to be partitioned into basins of attraction, e.g. [2]. The size of B($x$) is the average length of gradient walks terminating at $x$ [13]. In our analysis, B($x$) is the set of points such that a slow adaptive walker from a point in B($x$) terminates at $x$, and the cardinality of B($x$) is its size. This definition of basin of attraction and its size is comparable to that in [11] and [17]; except that our basins are formed with different step sizes. The basin of attraction for problems with a unique global optimum such as the NK problems is the entire search space or all points in the analyzed sample. We include $x$ in B($x$).

The distribution of basin sizes is an important performance factor for optimization heuristics. In particular is the correlation between basin size and fitness of the local optimum [13]. In their LON study of NK problems, both [11] and [17] report a strong positive correlation between local optimum fitness and basin size. We too observe significant (p-values $\leq$ 0.05) positive correlations from our study of the basins of attraction in all three sample types (Table 2 & Fig. 8). Thus slow adaptive walks can be a feasible method of exploring NK search spaces.

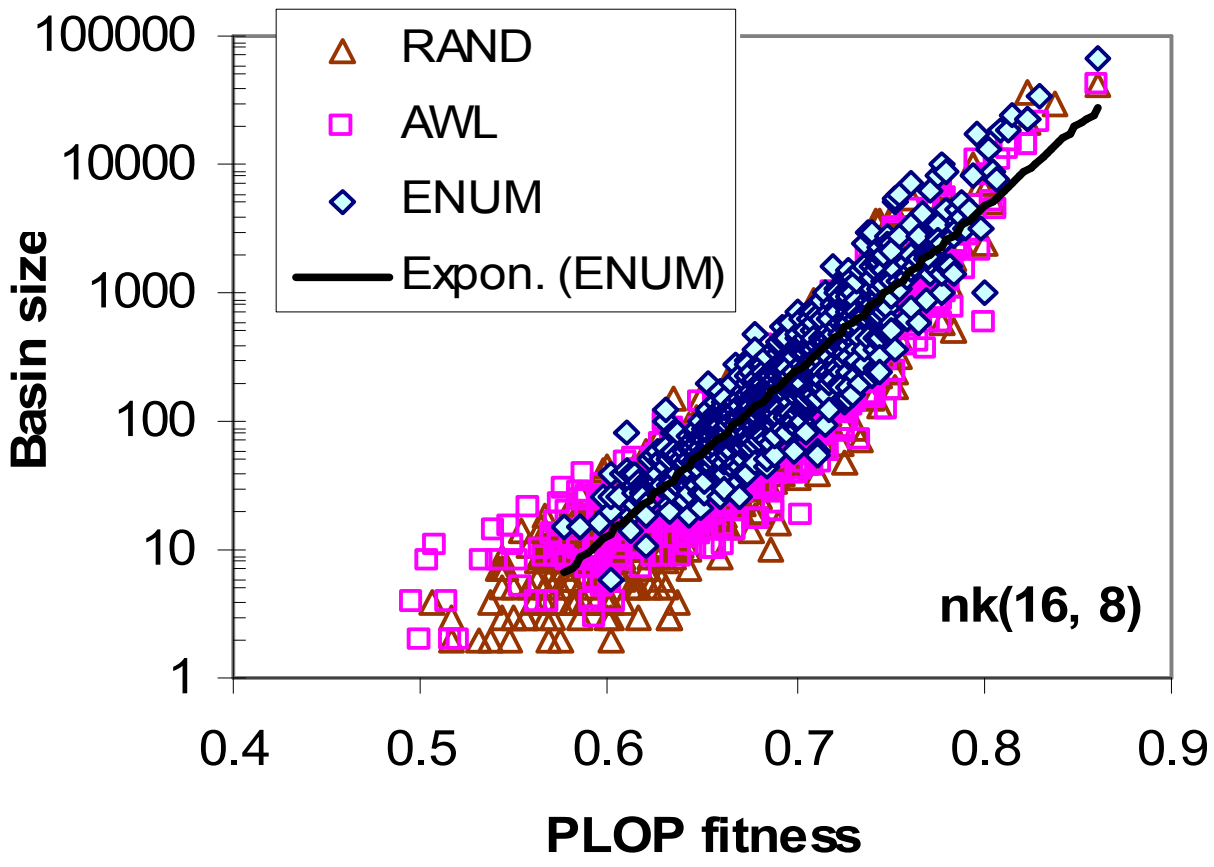

**Fig. 8 Scatter plots for a run chosen at random per NK problem. Only results for nk(16, 8) are shown.**

**Table 2 Average Spearman correlation coefficient ± standard deviation. All 30 coefficients were significant (p-value ≤ 0.05).**

| NK | ENUM | AWL | RAND |
|---|---|---|---|
| (16, 4) | 0.8968 ± 0.04455229 | 0.9223 ± 0.03609376 | 0.8761 ± 0.02046371 |
| (16, 8) | 0.9062 ± 0.01046025 | 0.9092 ± 0.009234367 | 0.8871 ± 0.01117023 |
| (16, 12) | 0.8840 ± 0.00615435 | 0.8801 ± 0.006299658 | 0.8737 ± 0.007595923 |

As K increases, AWL samples have smaller Mean Absolute Difference and Root Mean Squared Difference than RAND samples (Table 3). This means that AWL samples produce PLOP fitness basin size correlation coefficients closer in value to those produced by ENUM than RAND samples.

**Table 3 Differences between significant correlation coefficients**

| NK 16 | Mean absolute difference | | Root mean squared difference | |
|---|---|---|---|---|
| | ENUM, AWL | ENUM, RAND | ENUM, AWL | ENUM, RAND |
| 4 | 0.045440 | 0.039609 | 0.064070 | 0.051528 |
| 8 | 0.009811 | 0.019670 | 0.011655 | 0.023165 |
| 12 | 0.005533 | 0.010901 | 0.006850 | 0.012996 |

## 3.2 Step Size Barrier

To hop from one local optimum to a fitter local optimum involves crossing a barrier of some kind. Commonly (due to the fixed search radius paradigm), such barriers are quantified in terms of the minimum amount of fitness that needs to be sacrificed. Points at which such fitness barriers become possible are saddle points. The set of local optima and their saddle points can be organized into a hierarchical structure called a barrier tree [4, 13]. Alternatively, (with variable radius search such as the slow adaptive walker), distance may be used to quantify barriers. [10] alluded to the direct benefit of knowing distance barriers.

For a pair of PLOPs (*x, y*), the *step size barrier sb* for *y* is the minimum of the set of maximum step sizes in all slow adaptive walks from *x* to *y*. A step size barrier may not be found between all pairs of PLOPs. But when found, a step size barrier *sb* between a pair of PLOPs (*x*, *y*) forms a directed edge from *x* to *y* with weight *sb* in a *temperature network*. Nodes of a temperature network are PLOPs, and for samples with a unique fittest point, form a connected component (minimally, the fittest point forms the articulation node). A number of important analyses can be performed on a temperature network, e.g. pathways to the fittest point or the global optimum.

Presently, we focus on *sb* values in temperature networks formed from ENUM, AWL and RAND samples. For each run per NK problem, we take the mode (most frequently occurring) *sb* averaged over all nodes in a temperature network (excluding source nodes), and the mode *sb* for the fittest point. When compared with ENUM, the all nodes average mode *sb* (Table 4) and fittest node mode *sb* (Table 5) values produced by RAND samples show larger deviations than those produced from AWL samples. The differences are larger for fittest node mode *sb* (Table 5). Hence AWL samples produce more representative temperature networks than RAND samples.

**Table 4 Mean differences between all nodes average mode *sb***

| NK 16 | Mean absolute difference | | Root mean squared difference | |
|---|---|---|---|---|
| | ENUM, AWL | ENUM, RAND | ENUM, AWL | ENUM, RAND |
| 4 | 0.102506 | 0.134966 | 0.149345 | 0.176484 |
| 8 | 0.049553 | 0.082868 | 0.054650 | 0.091024 |
| 12 | 0.011757 | 0.019899 | 0.014252 | 0.023477 |

**Table 5 Mean differences between fittest node mode *sb***

| NK 16 | Mean absolute difference | | Root mean squared difference | |
|---|---|---|---|---|
| | ENUM, AWL | ENUM, RAND | ENUM, AWL | ENUM, RAND |
| 4 | 0.4 | 1.2 | 1.316561 | 2.097618 |
| 8 | 0.166667 | 1.7 | 0.408248 | 2.915476 |
| 12 | 0.466667 | 0.633333 | 1.064581 | 1.110555 |

## 3.3 Basin Overlap Network

To understand how basins of attraction are organized relative to one another in a search space, we construct a *basin overlap network* whose *V* nodes are PLOPs and *E* edges denote non-empty intersections between basins of attraction. Edges due to transitive subset relations are removed. Specifically, a directed edge (*x*, *y*) in a basin overlap network implies $x \neq y$, *x* is less fit than *y*, $B(x) \cap B(y) \neq \emptyset$ and if $B(x) \subset B(y)$ then $B(y) \setminus B(x)$ is minimal.

Although the edges in a basin overlap network are directed, it is still meaningful (since the less than relation is not symmetric) to treat them as undirected when computing link density, and clustering coefficient [20]. Hence the number of all possible edges between *v* nodes is $v(v-1)/2$. The degree of a node is the sum of its in degree and out degree. Mean degree $\langle k \rangle$ is $2E/V$ and link density is $\frac{2E}{V(V-1)}$ or $\frac{<k>}{V-1}$. Average path length is computed by following the direction of edges, and so it represents the number of edges to traverse on average from a PLOP to a fitter PLOP. To reduce computation time, average path length is computed for at most 1000 nodes chosen uniformly at random. These network statistics averaged over 30 basin overlap networks for each NK problem are presented in Fig. 9.

We discuss the ENUM results first. Both link density and clustering drop with increase in network size (number of PLOPs). However, for all network sizes tested, the clustering coefficient remains significantly larger than link density, which we use as proxy for the level of clustering expected of a random graph of the same size which estimates at $\langle k \rangle / V$. The high clustering levels cannot be attributed to transitive subset relations since these have been removed. The high clustering levels which decrease with increase in K are consistent with results for LONs on NK problems [11]. The average path length increases with network size. This is not surprising given that the basin overlap networks grow sparser as they get larger. However, the rate of increase shows deceleration, which on the one hand, could either be due to our computation method, or indicate that basin overlap networks are small worlds. The latter observation would be consistent with results reported for LONs on NK problems [11], and also for inherent networks of atomic clusters [3].

Node degree distribution is another network attribute used to characterize search spaces. A scale-free degree distribution has been linked to funnel shaped fitness landscapes which are deemed search easy [2]. Scale-free degree distributions have been observed in inherent networks of atomic clusters [3], but not in LONs of NK problems [11]. Our basin overlap networks for the NK problems are also not scale-free. At best, their degree distributions can be described as exponential (Fig. 10). The degree coefficient of variation (standard deviation to mean degree ratio) remains below 1.0 (Fig. 11 top).

Nonetheless, we find that by excluding some edges from the basin overlap network, a more scale-free like degree distribution can be obtained (Fig. 10). An edge (*x*, *y*) is excluded if *x* and *y* are not part of any slow adaptive walk. Excluding such edges increased the degree coefficient of variation to near or above 1.0 (Fig. 11 bottom) which is indicative of a broad heterogeneous degree distribution. It can be argued that this only exposes the degree distribution of the underlying network of slow adaptive walks which is scale-free (although a subnetwork of a scale-free network need not be scale-free [15]). But we do not see this as a deception. On the contrary, by restricting the search space dynamics to hopping from PLOP to fitter PLOP, we believe the exclusion of such edges brings the basin overlap network closer to inherent networks which so far has been constructed for systems where nearly all the minima and transition states can be located [2, 3], and conformational space networks which are constructed by sampling the landscape using long molecular dynamics simulation at a temperature near to the folding transition thereby increasing the likelihood of locating the global optimum which is suspected to strongly influence the appearance of the scale-free phenomenon [12]. From this, we take that degree distribution of a fitness landscape can indicate when a search has become easy which makes it a possible fitness function for search algorithms. It is also worth pointing out that RAND samples produced degree distributions similar in kind to ENUM and AWL samples, albeit it appears to be less differentiating between the NK problems (Fig. 10).

Curiously, the degree coefficient of variation of basin overlap networks increases with network size (Fig. 11). On the one hand, a heterogeneous degree distribution is easier to obtain in a large network with low link density than in a small network with high link density. So this makes sense in the general network sense. On the other hand, this implies that search becomes easier as the number of local optima increases. This runs contrary to theory. But perhaps we are giving the connection between degree distribution and search difficulty too broad an interpretation and comparisons between degree distribution types should only be made holding the number of local optima constant.

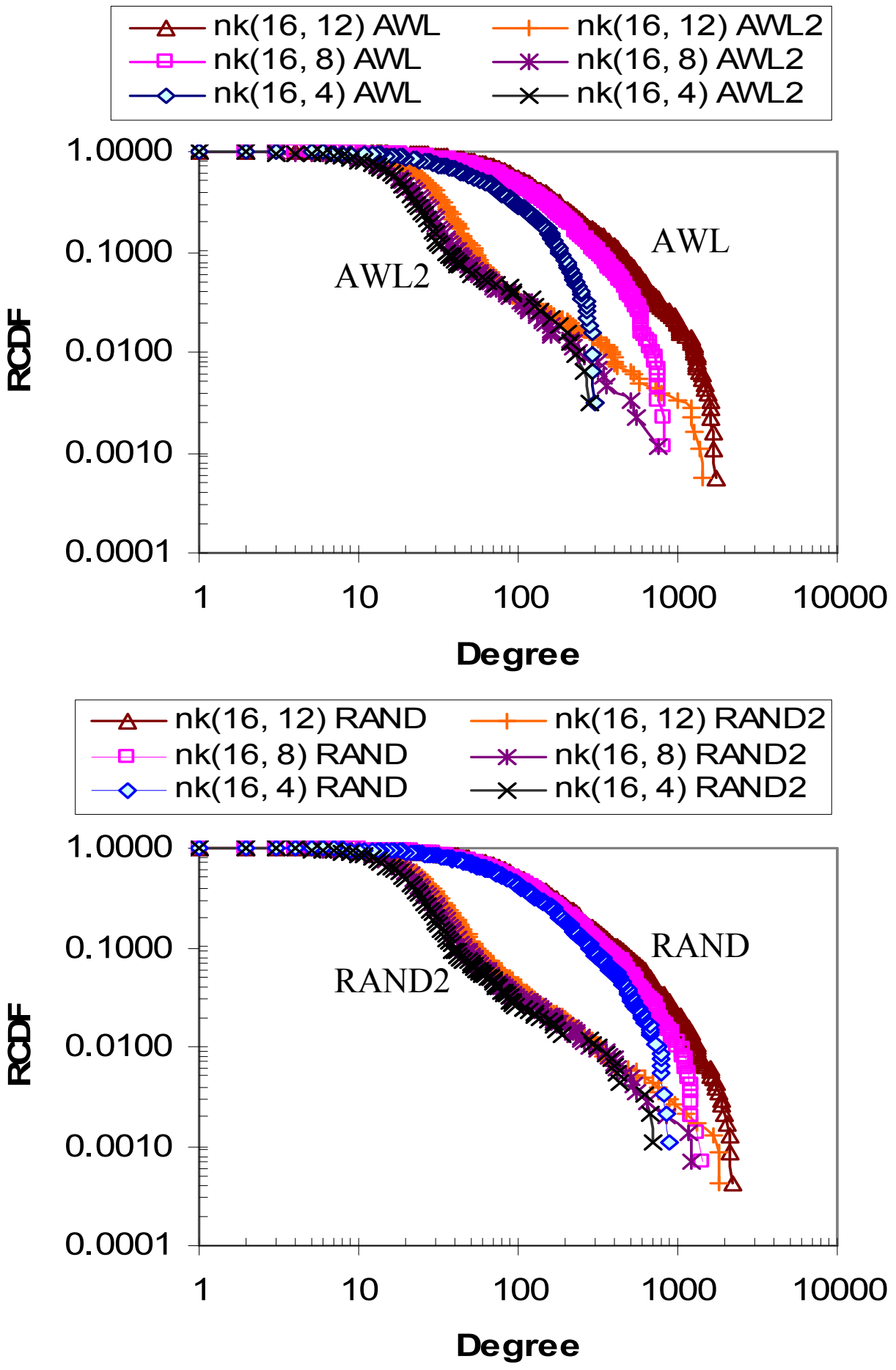

**Fig. 10 Reversed cumulative probability degree distributions chosen at random. Labels ending in '2' denote plots for networks with excluded edges.**

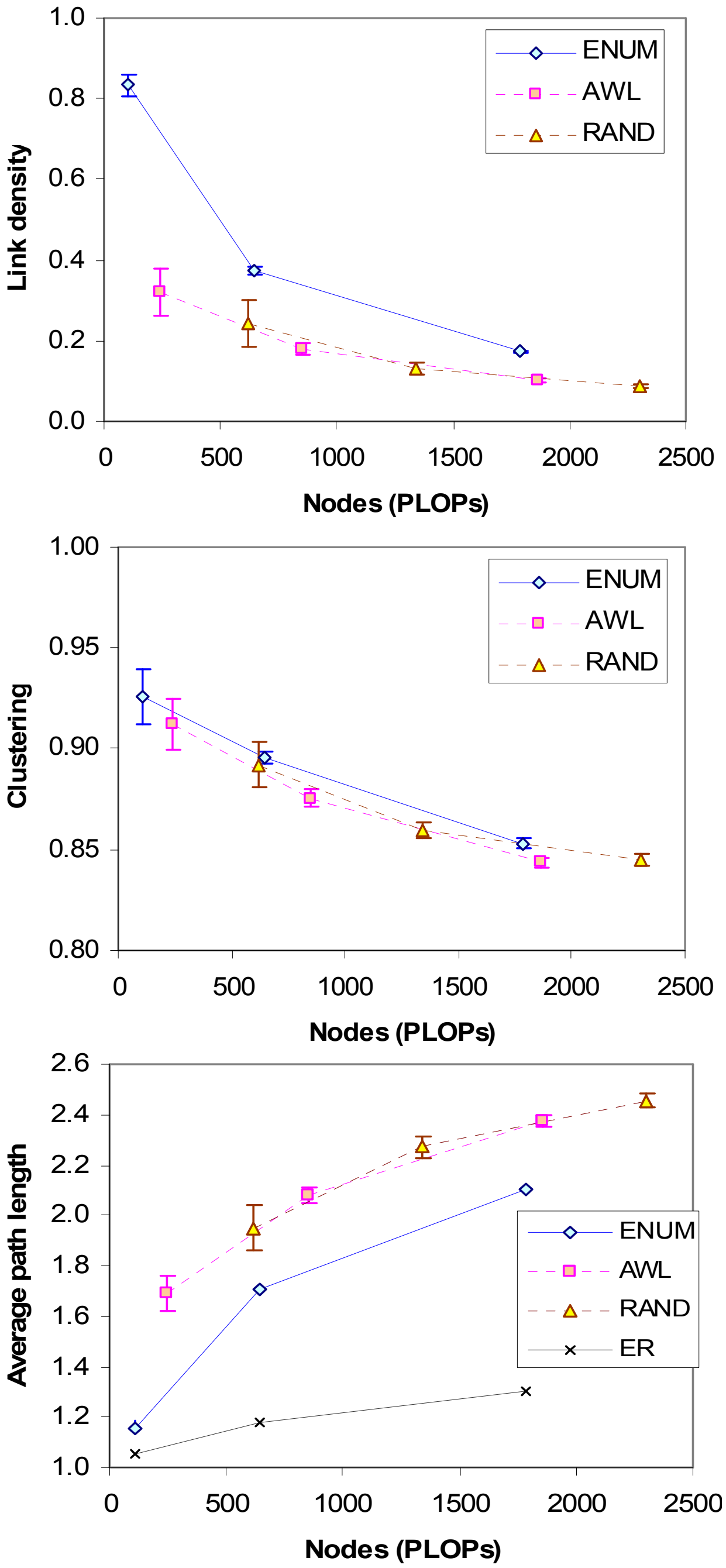

**Fig. 9 Network statistics averaged over 30 basin overlap networks for each NK problem. ER refers to the average path length expected of an Erdos-Renyi random graph ln N/ ln ⟨k⟩.**

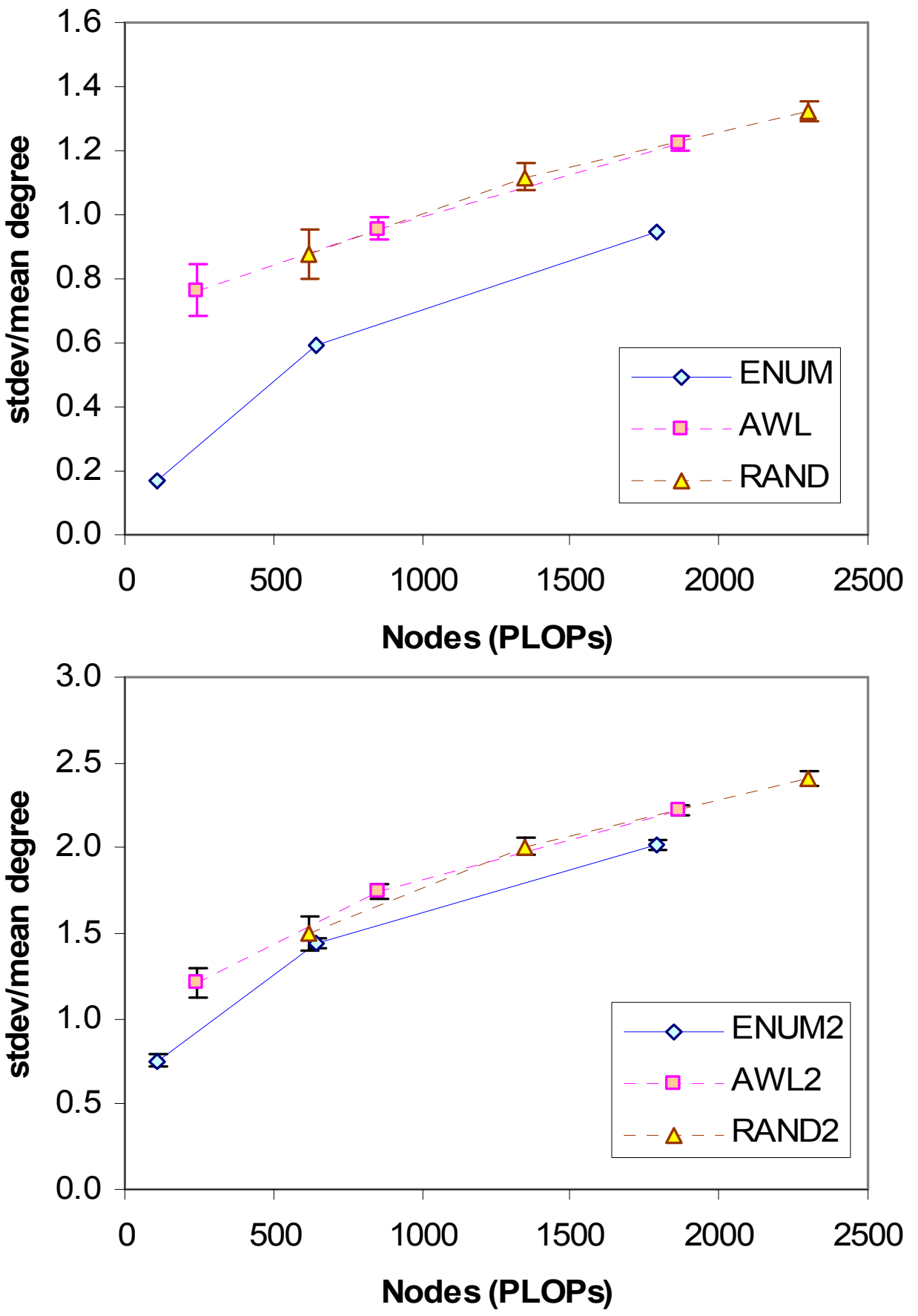

**Fig. 11 Degree coefficient of variation (standard deviation / mean) averaged over 30 basin overlap networks for each NK problem. Labels ending in '2' denote plots for networks with excluded edges. Error bars indicate 95% confidence interval.**

## 4. SENSITIVITY TO SLOWNESS

Finding a nearest fitter point is computationally expensive. In this section we test los sensitivity to this requirement under three circumstances: *dyna*, *rand* and *combi*. In *dyna*, the set of neighbors for each point is pre-computed using a sample of 1000 fitter points chosen uniformly at random for each point. In *rand*, the set of neighboring points are also computed using a sample of 1000 fitter points chosen uniformly at random for each point, but the search is carried out as the walks are made and at each time a point is reached. Therefore, *rand* does more sampling than *dyna*, but the sampling results are independent of each other. *combi* uses *rand* style sampling, but a memory of the best sampling results for each point is kept. So the more frequently a point gets visited by the slow adaptive walker, the more accurate its neighboring set of points is expected to be.

We compare PLOPs produced from *dyna*, *rand* and *combi* walks against PLOPs produced from walks using exhaustive search. Define *est* as the set of PLOPs produced from *dyna*, *rand* or *combi* walks, and *act* as the set of PLOPs produced from walks using exhaustive search. False positive (fp) is $\frac{|est \setminus act|}{|est|}$, false negative (fn) is $\frac{|act \setminus est|}{|act|}$, and overlap is $\frac{|est \cap act|}{|est \cup act|}$. Results presented in Fig. 12 are averaged over the 30 runs for each NK problem. Overall, particularly with larger K problems, *combi* produced the best results, with relatively lower levels of false positives and false negatives, and relatively higher overlap. However, the results are still very poor, with maximum overlap of about 0.15 only (K=12).

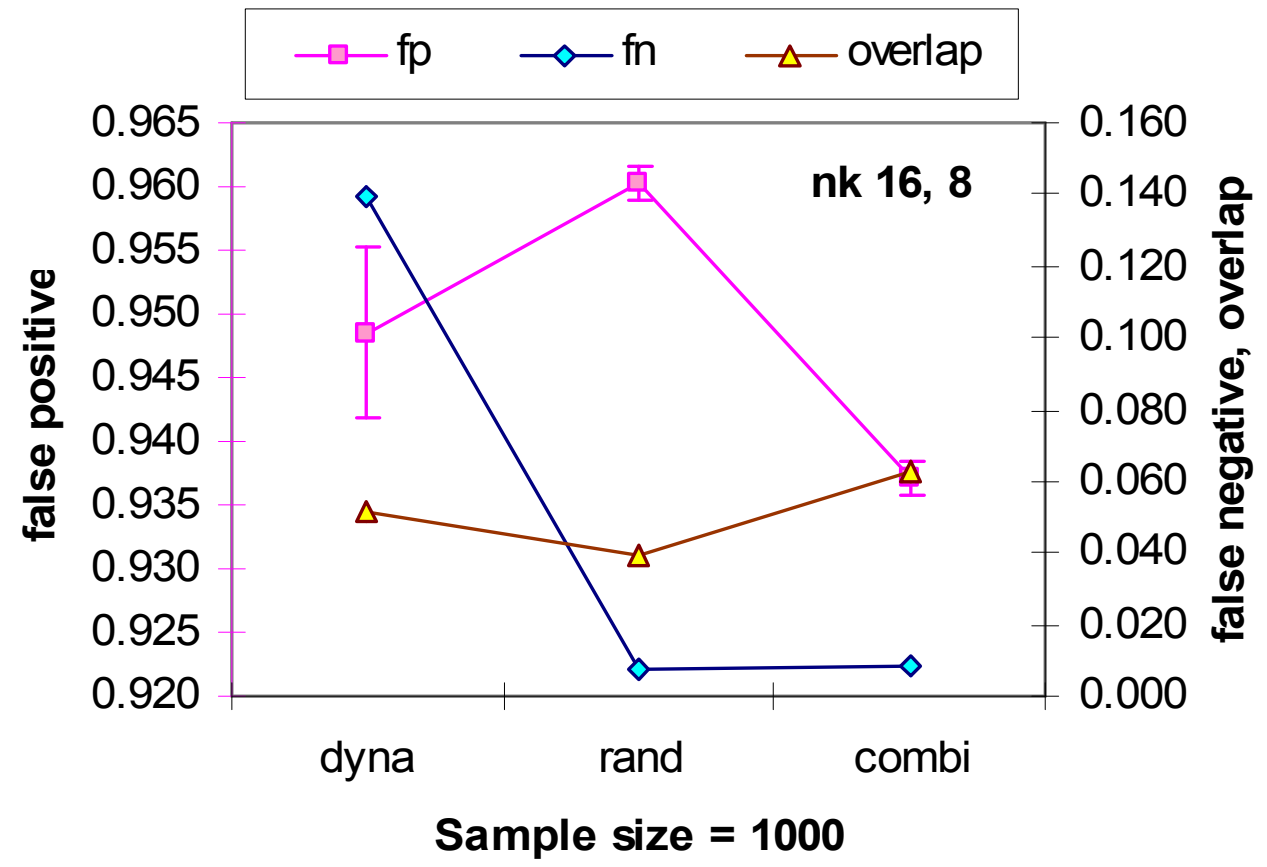

**Fig. 12 Results of three sampling strategies for finding nearest fitter points. Error bars indicate 95% confidence interval. Only results for nk(16, 8) are shown.**

But the results are interesting in themselves and reveal the importance of slowness of walks. Due to insufficient sampling, *dyna* has higher false positive and false negative rates. When sampling is increased (as it is in *rand*), false negatives drop drastically (more actual PLOPs in the sample are caught). However, more independent sampling introduces the possibility of more variation in the out step size, which on the one hand is a factor in reducing false negatives, but at the same time, also increases false positives (a wider net catches both what is wanted and what is not). By having a memory (to remember the best set of neighboring points found for a point so far), *combi* like *rand* does more sampling than *dyna*, but unlike *rand*, *combi* tames the variability of the out step size. From the results, this strategy is able to maintain significantly lower false negatives and false positives, and as a result achieve significantly higher overlap.

## 5. DIFFERENT TEST PROBLEM

Due to the practical issues with AWL sampling mentioned in section 2.3, and to add variation to the test set, we use the HIFFC (continuous HIFF) and HIFFM [21] problems. The minimum and maximum fitness values for these two test problems can be calculated directly, and we did not encounter problems with holes in the fitness spectrum for N=16. Unlike the NK problems, both the organization and weight of links in these two test problems are non-random; and they each have two global optima.

We use the same parameter values (section 2.3) for AWL sampling except fitness bins are in increments of 2 from 4 to 16 for HIFFC, and from 2 to 16 for HIFFM. With these parameters, the average sample size for both HIFFC and HIFFM covered about 60% of the search space. AWL samples contain significantly more locally optimal points (points with *plef* = 1.0) than RAND samples. This occurs despite RAND sampling over 50% of the search space. When a locally optimum point is present in a sample, it is detected 100% of the time in AWL samples; the *plef* LO detection rate is slightly but significantly lower in RAND samples. Further, PLOPs are overestimated in RAND samples to the extent that HIFFM ends up with significantly more local optima than HIFFC (Fig. 13), which is not the actual case. When N=16, the number of local optima, i.e. *plef*=1.0 points, for HIFFC and HIFFM are 256 and 122 respectively. In contrast, AWL samples suffer much less from overestimation errors, and the PLOPs detection algorithm correctly estimates fewer local optima for HIFFM than HIFFC.

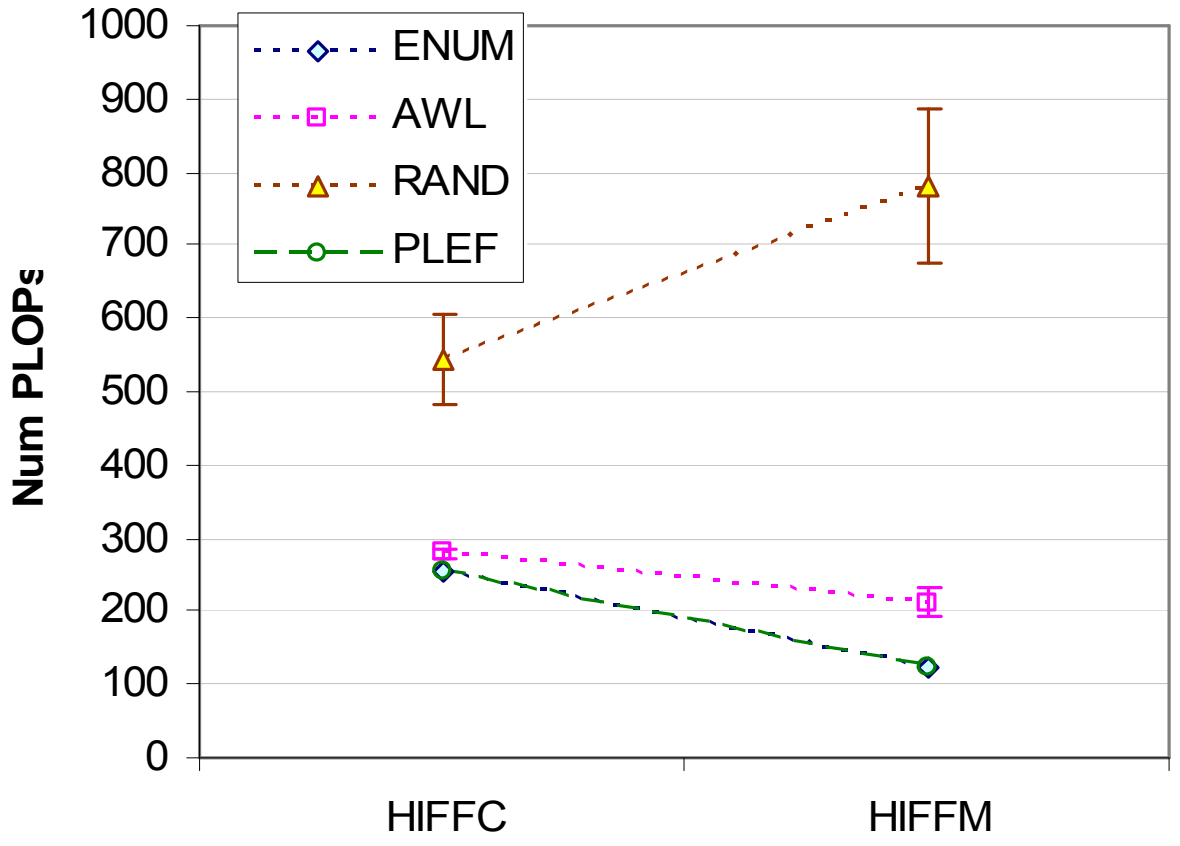

**Fig. 13 Number of PLOPs detected in a sample averaged over all samples/instances per problem for N=16. Error bars indicate the 95% confidence interval for the average. PLEF is the actual number of points in the entire search space with *plef* = 1.0 average over all instances per problem.**

Analysis similar to those conducted for the NK problems were repeated on the HIFFC and HIFFM problems. The results are reported in arXiv:1112.5980. Overall, the results favor the use of AWL over RAND samples

# 6. CONCLUSION

We proposed a local optima detection algorithm base on step sizes taken by a slow adaptive walker and showed that it works well to detect local optima present in a sample (Fig. 6).

We also explored the use of Wang-Landau sampling of search spaces and found that overall AWL samples produced better results than RAND samples. Specifically: (i) AWL samples contain significantly more local optimal points, i.e. points with *plef* = 1.0, than RAND samples (Fig. 5); (ii) the PLOP detection algorithm produced significantly fewer false positives with AWL samples than with RAND samples (Fig. 7); (iii) AWL samples produce PLOP fitness basin size correlation coefficients closer in value to those produced by ENUM than RAND samples (Tables 2 & 3); and (iv) AWL samples produce more representative temperature networks than RAND samples (Tables 4 and 5).

Our work on basin overlap networks led us to some surprising conclusions (Fig. 11) and doubts about the value of *qualitative* network characterization of fitness landscapes. However, this could also be due to the ability of slow adaptive walks to reach a fittest point in a sample.

We are working to develop these techniques for larger search spaces.

# 7. ACKNOWLEDGEMENTS

My thanks to Dr. P. Grogono for kindly arranging the use of computer resources at Concordia University, Montreal, Quebec. This work is self-funded.